\documentclass[journal]{IEEEtran}
\usepackage{amsmath,amsfonts}
\usepackage{algorithmic}
\usepackage{array}
\usepackage[caption=false,font=normalsize,labelfont=sf,textfont=sf]{subfig}
\usepackage{textcomp}
\usepackage{stfloats}
\usepackage{url}
\usepackage{verbatim}
\usepackage{graphicx}
\def\BibTeX{{\rm B\kern-.05em{\sc i\kern-.025em b}\kern-.08em
    T\kern-.1667em\lower.7ex\hbox{E}\kern-.125emX}}
\usepackage{balance}

\usepackage[numbers]{natbib}
\usepackage{multicol}
\usepackage{multirow} 
\usepackage{bm}
\usepackage{booktabs}
\usepackage[table]{xcolor}

\usepackage{color}

\newcommand{\etal}{\textit{et al}.}

\begin{document}
\title{Refining 3D Point Cloud Normal Estimation via Sample Selection}

\author{
~\IEEEauthorblockN{Jun Zhou, Yaoshun Li, Hongchen Tan, Mingjie Wang, Nannan Li, Xiuping Liu} \\
\thanks{The authors would like to thank the High Performance Computing Center of Dalian Maritime University for providing the computing resources. This research was supported in part by the Natural Science Foundation of China under Grants 62002040, 61976040, and 62201020, in part by China Postdoctoral Science Foundation 2021M690501, in part by the Science Foundation of Zhejiang Sci-Tech University under Grant number 22062338-Y, and in part by Beijing Postdoctoral Science Foundation under Grant number 2022-ZZ-069. (Corresponding author: Jun Zhou.)}
\thanks{J. Zhou,Y. Li and N. Li are with the School of Information Science and Technology, Dalian Maritime University, Dalian, China (E-mail: zj.9004@gmail.com, lys96@dlmu.edu.cn, nannanli@dlmu.edu.cn). }
\thanks{H. Tan is with the Institute of Artificial Intelligence,  Beijing University of Technology, Beijing , China (E-mail: tanhongchenphd@bjut.edu.cn). }
\thanks{M. Wang is with the School of Science, Zhejiang Sci-Tech University, Zhe Jiang, China (E-mail: mingjiew@zstu.edu.cn). }
\thanks{X. Liu is with the School of Mathematical Sciences, Dalian University of Technology, Dalian 116024, China. (E-mail: xpLiu@dlut.edu.cn). }
}


\markboth{Journal of \LaTeX\ Class Files,~Vol.~18, No.~9, September~2020}%
{How to Use the IEEEtran \LaTeX \ Templates}

\maketitle

\begin{abstract}
In recent years, point cloud normal estimation, as a classical and foundational algorithm, has garnered extensive attention in the field of 3D geometric processing. Despite the remarkable performance achieved by current Neural Network-based methods, their robustness is still influenced by the quality of training data and the models' performance. In this study, we designed a fundamental framework for normal estimation, enhancing existing model through the incorporation of global information and various constraint mechanisms. Additionally, we employed a confidence-based strategy to select the reasonable samples for fair and robust network training. The introduced sample confidence can be integrated into the loss function to balance the influence of different samples on model training. Finally, we utilized existing orientation methods to correct estimated non-oriented normals, achieving state-of-the-art performance in both oriented and non-oriented tasks. Extensive experimental results demonstrate that our method works well on the widely used benchmarks.

\end{abstract}

\begin{IEEEkeywords}
Normal estimation, Sample selection, Robust training
\end{IEEEkeywords}

\section{Introduction}
\IEEEPARstart{P}oint cloud normal estimation is a pivotal task in computer graphics. Due to the inherent unordered and non-uniform distribution of point clouds, normals serve as fundamental features that provide valuable additional information. Effective methods for normal estimation have proven valuable across various downstream applications such as odometry and mapping~\cite{vizzo2021poisson,guo2022loam}, 3D reconstruction~\cite{berger2014state,kazhdan2006poisson,hashimoto2019normal,kazhdan2013screened,huang2023neural}, point cloud denoising~\cite{zhang2020pointfilter}, and semantic segmentation~\cite{grilli2017review,che2018multi}. 

In recent years, deep learning methods have gained considerable attention for normal estimation task. In contrast to traditional approaches such as PCA-based methods~\cite{hoppe1992surface} and jets~\cite{cazals2005estimating},  deep learning techniques demonstrate promise in handling diverse data and exhibiting robustness to noise without requiring extensive parameter tuning. Typically, these methods adhere to a unified paradigm: they sample local neighborhood patches as input, use a neural network to extract features, and employ a regressor to estimate normals. This regressor can either directly estimate local normals or use surface fitting techniques. Training data is sourced from PCPNet~\cite{guerrero2018pcpnet} dataset. Algorithms developed based on this paradigm have achieved outstanding performance. Furthermore, thanks to their generalization capabilities, deep learning methods can be directly applied to various real scanned datasets, such as Semantic3D~\cite{hackel2017semantic3d}, SceneNN~\cite{hua2016scenenn}, and NYU depth v2~\cite{silberman2012indoor}.

\begin{figure*}[ht]
    \centering
    \includegraphics[width=0.79\textwidth]{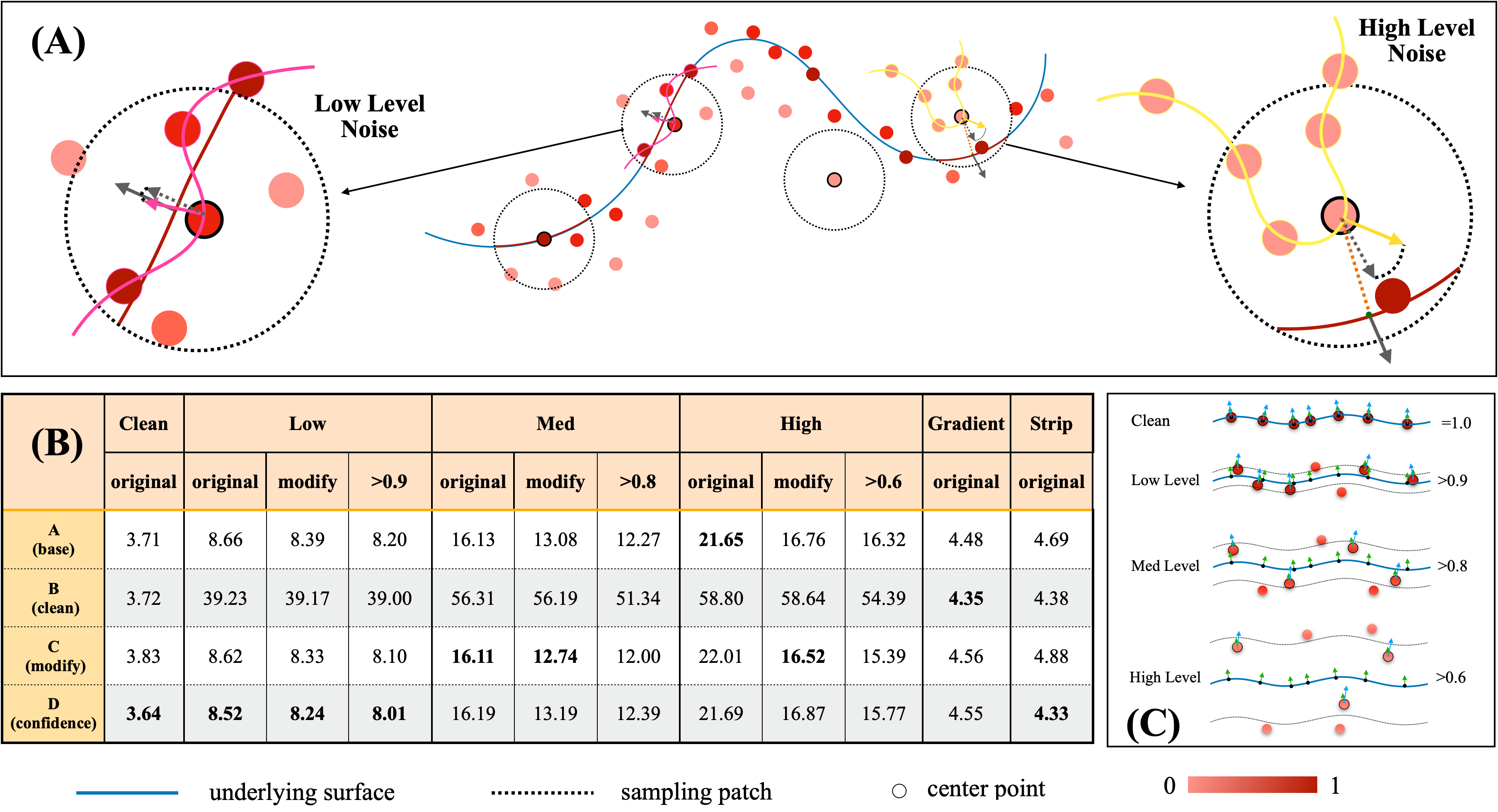}
    \caption{Explanation of how corrupt noise samples affect network training. (A) The top partillustrates the differences between low-level and high-level patches regarding their underlying surfaces. In high noise scenarios, local patches may lack clear surface patterns. Additionally, compared to patches with low noise levels, those with high noise may show notable differences between the normals of query points and those of the nearest points on the surface. These instances, known as corrupt samples, can weaken the model's robustness during training. (B) The table outlines four training strategies: A utilizes the entire PCPNET dataset, B exclusively uses clean data, C corrects normals from each point's underlying surface, and D employs our confidence-based training method. It shows that relying solely on clean data and corrected normals isn't ideal due to insufficient training data, resulting in reduced model robustness. (C) This part illustrates how the sampling range varies based on confidence values, which decrease as the noise scale increases.}
    \label{fig:1}
\end{figure*}

Benefiting from the feature extraction capabilities of neural networks, learning-based normal estimation algorithms, particularly with the current SHS-Net~\cite{li2023shs}, have achieved outstanding performance. However, existing methods still rely on the PCPNet~\cite{guerrero2018pcpnet} dataset for training, which contains various scales of noise. This diversity can enhance the network's generalization capability. Nevertheless, as the scale of model noise increases, there is a higher probability that sampled training samples will deviate further from the underlying clean surface, as illustrated in the top part of Fig.~\ref{fig:1}. Consequently, both the patterns of sampled patch and their corresponding ground truth will deviate from the ideal scenario, introducing ambiguous information into the training process. This limitation can restrict the algorithm's performance, causing trained models to produce offset results in noise-free data. As depicted in the table portion of Fig.~\ref{fig:1}, we evaluate a single-scale simple normal estimation model. Training solely with clean data or the entire dataset with corrected data adversely impacts the test results of clean data. However, applying confidence value constraints weakens the impact of corrupted data on noise, thereby enhancing model robustness.

To tackle this issue, we devised an evaluation process for each training sample, assigning individual confidence weights to them. These weights can be incorporated into the loss function, enabling the sample selection can be implemented by soft constraint way. This soft constraint approach offers advantages over direct ground truth correction. It not only addresses inconsistencies in ground truth data linked to high noise but also acknowledges inherent irregularities in noisy data itself. Consequently, this method aids in mitigating model biases. Extensive implementation analyses have shown its effectiveness in maintaining model performance on low noise datasets while also yielding satisfactory results on datasets with substantial noise levels. In addition, regarding the architecture of the normal estimation network, we also validated the effectiveness of multiple modules. By introducing global information and incorporating constraints such as the QSTN module, z loss, and local loss, we enhanced the model's performance. Finally, leveraging existing neural gradient techniques for correction, our approach demonstrates competitiveness in both unoriented normal estimation and consistent normal orientation tasks.


In summary, the contributions of this study are threefold:
\begin{itemize}
    \item A two-branch normal estimation neural network is proposed. This framework simultaneously captures local information of query points and their relative global context. By integrating multiple constraints, we construct a more robust baseline normal estimation model.
    \item Two methods for estimating confidence values of sampling patches are proposed. These values are then used as weights in the loss function to select reasonable samples and suppress corrupt ones. Compared to directly using corrected normals, our approach leads to a more robust normal estimation model.
    \item  We utilize existing orientation methods to update the normals estimated by our model, thereby obtaining normals that are consistent in orientation and precise in accuracy. Finally, we experimentally demonstrate that our method is able to estimate normals with high accuracy and achieves the state-of-the-art results in both unoriented and oriented normal estimation.
\end{itemize}

\section{Related Work}
In this section, we mainly review the task of unoriented normal estimation, which can be generally divided into two categories: traditional methods and learning-based methods.Then, to ensure overall completeness, we also briefly review the task of consistent normal orientation in the final subsection.

\subsection{Traditional Methods}
Normal estimation in point clouds is a well-established field in geometry processing. Principal Component Analysis (PCA)~\cite{hoppe1992surface} is the most renowned method, involving sampling fixed-size neighbors and using statistical algorithms to fit a local tangent plane. Variations like Moving Least Squares (MLS)~\cite{levin1998approximation}, truncated Taylor expansion fitting (n-jet)~\cite{cazals2005estimating}, and local spherical surface fitting~\cite{guennebaud2007algebraic} have also been proposed within this paradigm, introducing more advanced fitting strategies to mitigate the impact of noise levels and patch scales on algorithm accuracy. Additionally, robust statistics-based methods~\cite{fleishman2005robust,li2010robust,yoon2007surface,mederos2003robust,wang2013adaptive,wang2013consolidation} have been developed to estimate local patches reasonably and alleviate the influence of patch anisotropy on accuracy. However, these methods still struggle with oversmoothing sharp features and geometric details. Consequently, techniques based on Voronoi diagrams~\cite{alliez2007voronoi,amenta1998surface,merigot2010voronoi}, Hough transform~\cite{boulch2012fast}, and plane voting~\cite{zhang2018multi} have been proposed. While these methods offer strong theoretical guarantees for stability and accuracy, they often require tedious parameter tuning for different data models. In recent years, driven by the rapid development of deep learning technology, data-driven approaches for normal estimation have emerged, achieving remarkable results.

\subsection{Learning-based Methods}
As a pioneering work, HoughCNN~\cite{boulch2016deep} was the first to integrate deep learning models into normal estimation tasks. They utilized image information constructed from a transformed Hough space accumulator as input and employed Convolutional Neural Networks (CNNs) for estimating point cloud normals. This method demonstrated outstanding performance at the time and laid the foundation for the subsequent introduction of deep learning techniques based on point clouds. Following this, a series of more competitive algorithms emerged, including PCPNet~\cite{guerrero2018pcpnet} based on point cloud representation, Nesti-Net~\cite{ben2019nesti} based on 3D modified Fisher vector (3DmFV) representations and IterNet~\cite{lenssen2020deep} based on graph representation, are proposed. These frameworks notably improved the accuracy and efficiency of normal estimation algorithms. Particularly, PCPNet~\cite{guerrero2018pcpnet} garnered significant attention from researchers due to its simplicity and direct operation on point clouds. Drawing inspiration from this approach, various algorithms have been introduced to optimize normal estimation methods, broadly falling into two categories: deep surface fitting and direct regression-based methods. Inspired by this approach, various algorithms have emerged to enhance normal estimation methods, broadly divided into two categories: deep surface fitting and direct regression methods.

Deep surface fitting methods specifically combine deep neural networks with traditional fitting techniques to accurately estimate normals from point clouds. For instance, Cao~\etal~\cite{cao2021latent} propose a differentiable RANSAC-like module at the end of a point cloud-based neural network to predict a latent tangent plane. Lenssen~\etal~\cite{lenssen2020deep} utilize a learnable anisotropic kernel to iteratively fit the local tangent plane. Additionally, Refine-Net~\cite{zhou2022refine} explores refining the initial normal for each point by considering various representations. Although various efforts have been made by the aforementioned models to enhance local fitting performance, the assumption of a local first-order plane is not suitable for handling diverse patch styles. Therefore, DeepFit~\cite{ben2020deepfit} utilizes a neural network to learn point-wise weights, enabling weighted least-squares surface fitting. Subsequently, several variations of DeepFit have been proposed to further bolster the robustness of the fitting process. Among them, Zhang~\etal~\cite{zhang2022geometry} employ pre-estimated weights to guide the network in learning, thereby improving weight estimation accuracy. AdaFit~\cite{zhu2021adafit} introduces a novel layer to aggregate features from multiple global scales and predicts point-wise offsets to enhance normal estimation accuracy. The offset strategy resembles that of prior work~\cite{zhou2023improvement}, albeit with the inclusion of offsets for the top-k points. Additionally, GraphFit~\cite{li2022graphfit} integrates graph convolution and adaptive fusion layers into the weight estimation network. Du~\etal~\cite{du2023rethinking} introduce two simple strategies, including a z-direction alignment loss and a learnable residual term, which significantly enhance DeepFit and its variants. Despite these advancements mitigating the algorithm's sensitivity to fitting order, they have not fundamentally resolved the issue. Consequently, deep surface fitting methods continue to grapple with challenges related to overfitting and underfitting.

In the early stages, direct regression models based on point cloud network frameworks, such as PCPNet~\cite{guerrero2018pcpnet} and its variants~\cite{zhou2020normal}, face limitations due to network performance issues and can not achieve outstanding performance. However, recent advancements in point cloud frameworks, like Transformer, have enabled direct regression methods to exhibit stronger capabilities. For instance,  Zhou~\etal~\cite{zhou2022fast} introduce a fast patch stitching method utilizing transformer modules for direct multi-normal regression within a patch. Similarly, MSECNet~\cite{xiu2023msecnet} introduce a more effective Multi-Scale Edge Conditioning stream framework, resulting in higher accuracy. Unlike this paradigm of direct multi-normal regression for a patch, another type of direct regression methods still follows the approach of PCPNet~\cite{guerrero2018pcpnet}, where only the normal of patch center is regressed. Recent works in this category include HSurf-Net~\cite{li2022hsurf} and SHS-Net~\cite{li2023shs}, which first transform point clouds into a hyperspace through local and global feature extractions and then conduct plane fitting in the constructed space. Additionally, NeAF~\cite{li2023neaf} infers an angle field around the ground truth normal to provide more information for learning from the input patch. Moreover, CMG-Net~\cite{wu2023cmg} identifies inaccurately annotated training samples and proposed the Chamfer Normal Distance to address this issue. However, the normal corrections introduced by the algorithm may not always be entirely reasonable. Even after correction, the normals of points far from the underlying surface in the presence of significant noise may still contain errors. Furthermore, this type of patch lacks sufficient underlying surface information, reducing the accuracy of the trained model for evaluating normals of low-noise point clouds. In this work, we propose a training strategy to select reasonable samples, significantly enhancing the performance of point cloud normal estimation algorithms.

\subsection{Consistent Normal Orientation}
Note that the normals estimated by previous methods lack consistent orientation, a crucial aspect that has been extensively studied. Classic methods~\cite{hoppe1992surface,konig2009consistent,schertler2017towards,seversky2011harmonic,wang2012variational,xu2018towards,jakob2019parallel}, inspired by the Minimum Spanning Tree (MST) algorithm, employ local diffusion strategies to achieve consistent normals for point clouds. Recently, ODP~\cite{metzer2021orienting} proposes utilizing a neural network to learn oriented normals within local neighborhoods, introducing a dipole propagation strategy for global consistency. However, lacking global information incorporation renders this method less robust. Then, some approaches~\cite{mello2003estimating,mullen2010signing,walder2005implicit,huang2019variational,alliez2007voronoi,katz2007direct,chen2010binary,xie2004surface,xiao2023point,xu2023globally} based on volumetric representation have been proposed to enhance orientation consistency. Current learning-based methods for consistent orientation estimation can be categorized into two types: direct regression using neural networks~\cite{li2023shs,wang2022deep} and those based on the Neural Gradient Function technique~\cite{li2023neural, li2024neuralgf}. Due to their superior capability in capturing global consistency, direct neural gradient methods enhance qualitative accuracy, though potentially compromising quantitative precision in normal regression. Hence, in this paper, we employ NeuralGF~\cite{li2024neuralgf} to refine estimated normals, ensuring their effective application in 3D reconstruction tasks.

\begin{figure*}[ht!]
    \centering
    \includegraphics[width=0.79\textwidth]{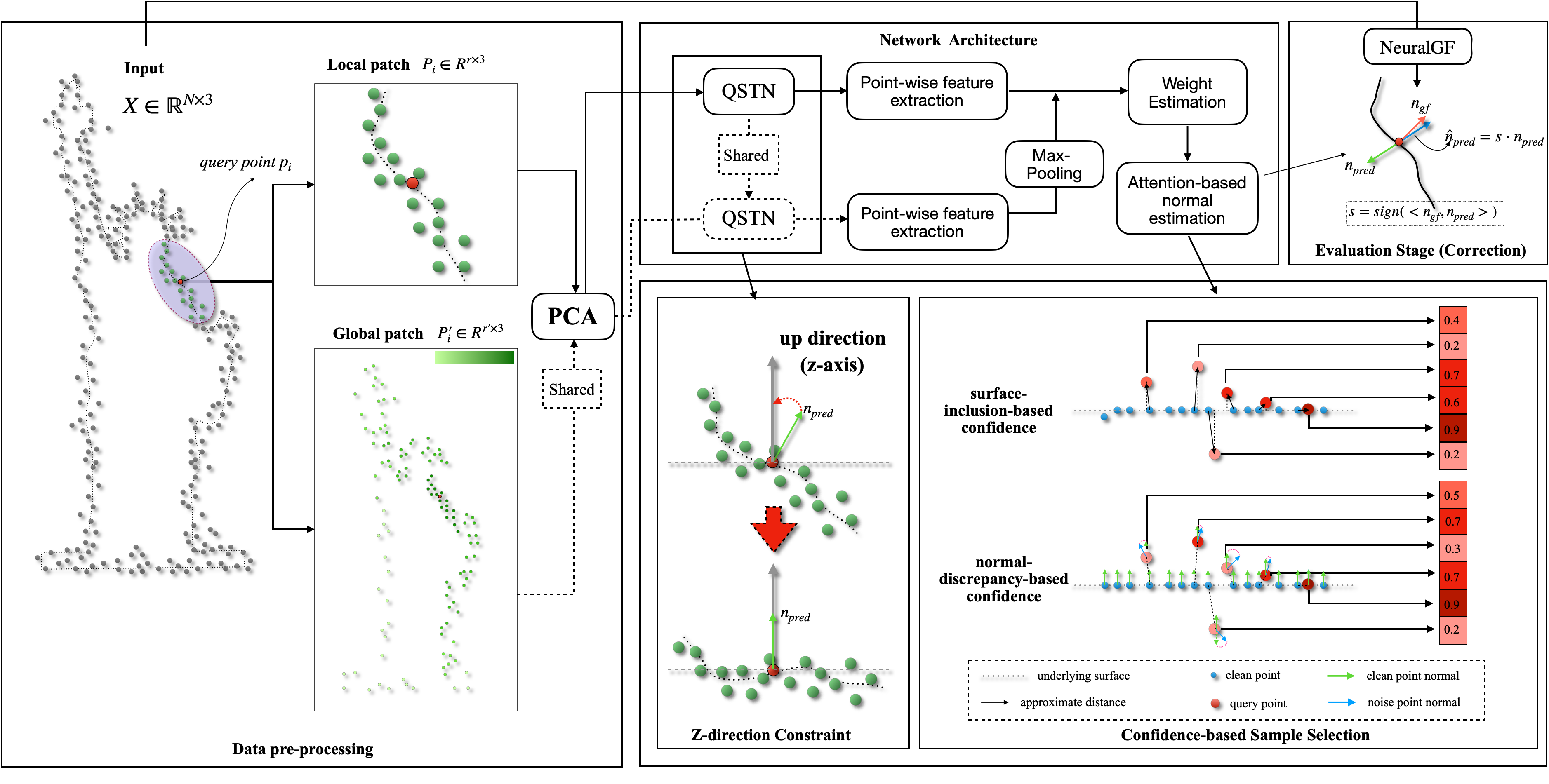}
    \caption{The learning pipeline of our method. Data preprocessing: local patches of query points and globally sampled patches based on probabilities are initialized after PAC processing. Network architecture: shared QSTN aligns multiple branch inputs, while NeuralGF method rectifies normal orientation. Sample selection: two strategies for estimating confidence values are presented, with the surface-based confidence assessing the distance of each point to the potential surface, and the normal-based confidence evaluating the disparity between the normals of each point and those of the potential surface.}
    \label{fig:2}
\end{figure*}

\section{Method}

\subsection{Overview}
The overall pipeline is depicted in Fig.~\ref{fig:2}. Similar to SHSNet~\cite{li2023shs}, a regression network is used to directly estimate the normal of the center point within a local patch. However, in contrast to SHSNet~\cite{li2023shs}, we enhance the network's performance by introducing a rotation module and additional constraint losses.
Furthermore, inspired by the integration of global information in SHSNet~\cite{li2023shs}, our approach incorporates globally sampled points as auxiliary information during normal regression. Detailed descriptions of the network architecture are provided in Sec.~\ref{sec:3.3}. In Sec.~\ref{sec:3.4}, we present two simple mechanisms for estimating the confidence of training samples based on position and normal differences with the underlying surface, respectively. Additionally, we introduce a confidence-based training strategy for sample selection, as described in Sec.~\ref{sec:3.5}. Finally, to address potential errors in normal orientation introduced by PCA techniques, we employ the current  popular method, namely, NeuralGF~\cite{li2024neuralgf} , to refine the orientation of the estimated normals (Sec.~\ref{sec:3.6}).

\subsection{Data pre-processing}
As shown in the left part of Fig.~\ref{fig:2}, given a 3D point clouds $X = \{p_{1},p_{2},\cdots,p_{L}\} \in R^{L\times 3}$, we first sample a local patch $P_{i} =\{p_{i,j}|p_{i,j}\in KNN(p_{i}),j=1,2,\cdots, r\}$ for each query point $p_{i}\in X$ using k-nearest neighbor (kNN) search. Additionally, we sample a global set $P'_{i}$ including $r'$ points for the query point $p_{i}$, representing a patch with globally distributed points on the shape surface. As in previous works~\cite{li2023shs, wang2022deep}, we introduce a probability-based sampling strategy to capture the local structure of the query point $p_{i}$ while retaining global information of the point cloud $X$. This strategy samples global points according to a density gradient that decreases with increasing distance from the query point $p_{i}$. The gradient of a point from $X$ can be calculated as follows:
\begin{equation}
    g(p_{j})  = \left\{
        \begin{array}{ll}
            \left[1-1.5\frac{\|p_{j}-p_{i}\|_{2}}{\max_{q\in X}\|q-p_{i}\|_{2}}\right]_{0.05}^{1} &\\
            1  &if~ j\in \mathcal{R}.
        \end{array}
    \right.
\end{equation}
where $[\cdot]_{0.05}^{1}$ indicates value clamping, and $\mathcal{R}$ is a random sample index set of point cloud $X$ with $r'$ items. Consequently, we can sample points according to probability distributions. We then use the classical PCA method to align the sampled patches, both local and global, as the initial input to our network.

\subsection{Architecture configuration}\label{sec:3.3}
As shown in the top part of Fig.~\ref{fig:2}, our network framework comprises two branches: one for encoding features from local patches and the other from the global point cloud, similar to the architecture proposed by SHS-Net~\cite{li2023shs}. However, unlike SHS-Net, both local patch $P_{i}$ and global point cloud inputs $P^{'}_{i}$ are initially passed through a shared QSTN module~\cite{du2023rethinking,guerrero2018pcpnet} to initialize the orientation of the input patches. As the orientation determined by the local patch is more accurate, we share this QSTN module estimated by local patches. Then, like SHS-Net~\cite{li2023shs}, we group the local features by k-NN and capture geometric information using MLP and max-pooling. For feature encoding, we also use a distance-based weight strategy to evaluate the importance of each point. To allow each point in the local patch to acquire global information, we use max-pooling and repetition operations to ensure the global latent code has the same dimension as the local latent code. The two codes are then fused by concatenation. Finally, through hierarchical information fusion and attention-weighted normal prediction, the final normal of the patch center $n_{i}^{pred}$ and the neighbors' normals are predicted.


\subsection{Confidence Estimation}\label{sec:3.4}
First, we identify two potential issues during the training stage: 1) unreliable annotated normals exist for some training samples , as shown in the bottom right corner of Fig.~\ref{fig:1}, and 2) noise patches that may lack sufficient underlying surface structure, as depicted in the upper part of Fig.~\ref{fig:1}. For this reason, we propose two confidence-based strategies, namely surface-inclusion-based and normal-discrepancy-based confidence, to evaluate the reliability of training samples. These strategies can effectively mitigate the impact of corrupted samples on the robustness of the model during the training process.

Specifically, the surface-inclusion-based strategy primarily involves evaluating the distance from each sampled patch's query point to the underlying surface. 
It is evident that points located far from the underlying surface are unlikely to effectively capture local surface structure information. Given a query point $p_{i} \in X$, our objective is to compute the distance from this point to the underlying surface, denoted as $d^{S}_{i} = D(p_{i},X)$. However, estimating the underlying surface for the model $X$ affected by noise is challenging. Therefore, we use the corresponding noise-free point cloud  $\hat{X}$ and assume it can represent the entire underlying surface. Based on this assumption, we compute the surface distance as  $d^{S}_{i} = D(p_{i},\hat{X})$. The noise-free point cloud $\hat{X}$ is represented as a point set $\hat{X} = \{\hat{p}_{1},\hat{p}_{2},\cdots,\hat{p}_{L}\}\in R^{L\times 3}$, and the distance $d^{S}_{i}$ can be approximately calculated as follows:
\begin{equation}
d^{S}_{i} = \min_{\hat{p}_{j}\in \hat{X}}\|p_{i}-\hat{p}_{j}\|_{2}.
\end{equation}
Consequently, the surface-inclusion-based confidence value can be calculated:
\begin{equation}
c^{S}_{i} = \exp(-\frac{d^{S}_{i}}{s \sigma_{S}}),
\end{equation}
where $s$ denotes the scale value of the given point cloud $X$, and $\sigma_{S}$ is set to 0.05 based on extensive experimentation. More details and comparative analyses are provided in the subsequent ablation experiments.

In addition to the surface-inclusion-based confidence estimation strategy, we also introduce a normal-discrepancy-based confidence estimation strategy. We assume that the true normal for each query point should be consistent with the normal near the underlying surface. While this assumption may not always hold, it helps filter out corrupted training samples. Specifically, for a query point $\hat{p}_{i}$ with an annotated normal vector $n_{i}\in R^{3}$, we search for the nearest point $\hat{p}_{i}$ on the noise-free point cloud $\hat{X}$. The point $\hat{p}_{i}$ approximates a surface point on $S$. Therefore, the for the query point $p_{i}$, the normal on the approximation surface point can be obtain as follows:
\begin{equation}
\hat{n}_{i} =  \{\hat{n}_{j} |\arg\min_{\hat{p}_{j}\in \hat{X}} \|p_{i}-\hat{p}_{j}\|\}. 
\end{equation}
Then, the difference between the normal of the query point and the normal near the surface point can be calculated as follows:
\begin{equation}
    d_{i}^{N} = \frac{\arccos (|{\langle \mathbf{n}_{i}, \hat{\mathbf{n}}_{i}}\rangle|)}{\pi/2},
\end{equation}
where ${\langle \cdot, \cdot}\rangle$ represents the inner product of two vectors and $\arccos (\cdot)$ function computes the inverse cosine, returning the angle in radians. Finally, the normal-discrepancy confidence value can be calculated as follows:
\begin{equation}
c^{N}_{i} = \exp(-\frac{d^{N}_{i}}{\sigma_{N}}),
\end{equation}
where $\sigma_{S}$ is set to 0.06 based on extensive experimentation.
  
\subsection{Loss function}\label{sec:3.5}
Firstly, similar to the approach in \cite{du2023rethinking}, we use the transformation regularization loss and the z-direction transformation loss to constrain the output rotation matrix
$R_{i} \in R^{3\times3}$ of the QSTN operation for the $i$-th sample of the point cloud $X$. The goal of these regularization losses is to achieve a rigid transformation, aligning the sampled patch vertically with the z-axis as much as possible, thereby minimizing extra degrees of freedom in the given patch $P_{i}$.These two loss functions are are defined as follows:
\begin{equation}
    L_{1} = \|I-R_{i}R_{i}^{T}\|^{2},
\end{equation}
\begin{equation}
    L_{2} = \|n_{i}R_{i}\times z\|
\end{equation}
where $I \in R^{3\times3}$ represents the identity matrix and $z = (0, 0, 1)$. 

In addition to the regularization loss, we also introduce the center loss, neighborhood consistency loss, and weight constraint loss. Unlike previous methods, we re-weight the loss term for the $i$-th sample using the corresponding confidence value $c_{i}$. Here, $c_{i}$ can be chosen either surface-inclusion-based confidence or normal-discrepancy-based confidence. Specifically, the re-weighted center loss for the patch $P_{i}$ involves minimizing the sine loss between the estimated normal $n^{pred}_{i}$ and the ground truth $n^{gt}_{i}$ as follows:
\begin{equation}
    L_{3} = c_{i}\|n^{pred}_{i}\times n^{gt}_{i}\|.
\end{equation}
The weight constraint loss~\cite{zhang2022geometry} incorporating our confidence value can be written as:
\begin{equation}
    L_{4} = c_{i} \frac{1}{M}\sum_{k=1}^{M}(w^{pred}_{i,k}-w^{gt}_{i,k})^{2},
\end{equation}
where $w^{pred}$ are the predicted weights for each point in the given patch, $M$ represents the cardinality of the downsampled patch, $w^{gt}_{i,k} = \exp(-(p_{i,k}\cdot n^{gt}_{i})^{2}/ \delta^{2})$ and $\delta = \max(0.05^{2},0.3\sum_{k=1}^{M} (p_{i,k}\cdot n^{gt}_{i})^{2}/M)$, where $p_{i,k}$ is a point in the downsampled patch of $P_{i}$.
Moreover, the neighborhood consistency loss emphasizes the importance of local points near the query center. The loss function for patch $P_{i}$ is expressed as:
\begin{equation}
    L_{5} = c_{i} \sum_{k=1}^{M} w^{pred}_{i,k} \|n^{pred}_{i,k}-n^{gt}_{i,k}\|^{2},
\end{equation}
where the neighborhood point normals $n^{pred}_{i,k}$ are predicted by our network. Therefore, the final loss function for the sample patch $P_{i}$ is defined as follows:
\begin{equation}
    L = \lambda_{1}L_{1}+\lambda_{2}L_{2}+\lambda_{3}L_{3}+\lambda_{4}L_{4}+\lambda_{5}L_{5}
\end{equation}
where $\lambda_{1}=0.1$,$\lambda_{2} = 0.5$, $\lambda_{3}=0.1$, $\lambda_{4} = 1$, and $\lambda_{5} = 0.25$ are weighting factors.

\subsection{Normal orientation correction}\label{sec:3.6}
As is well-known, direct regression-based normal estimation methods using neural networks often struggle to balance both orientation and accuracy in estimation. To address this problem, our study separates these aspects: the network focuses solely on regressing accuracy, while NeuralGF~\cite{li2024neuralgf} is introduced to estimate orientation of the entire input model using neural gradient functions. Details of this method can be found in the referenced work~\cite{li2024neuralgf}. With this approach, we can obtain the orientation $n^{gf}_{i}$for each query point of the given model. Then, as shown in the upper right part of Fig.~\ref{fig:2}, the orientation of the normal $n^{pred}_{i}$ can be corrected using the following formula:
\begin{equation}
    \hat{n}^{pred}_{i} = sign(n^{gf}_{i} \cdot n^{pred}_{i}) n^{pred}_{i},
\end{equation}
where $sgn(\cdot)$ is the signum function.

\begin{table*}[htp]
\caption{Comparison of the RMSE angle error for unoriented normal estimation of our method to classical geometric methods, and deep learning methods on datasets PCPNet and FamousShape.$*$ means the code is uncompleted.} 
\tiny
\centering
\begin{tabular}{l||cccc|cc|c||cccc|cc|c||cc|c}
\toprule
\multirow{3}{*}{\textbf{Category}}  & \multicolumn{7}{c||}{\textbf{PCPNet Dataset}} & \multicolumn{7}{c||}{\textbf{FamousShape Dataset}} &\multicolumn{3}{c}{\textbf{SceneNN Dataset}} \\
\cmidrule(r){2-18}
~ & \multicolumn{4}{c}{Noise} \vline &\multicolumn{2}{c}{Density} \vline &\multirow{2}{*}{Average} & \multicolumn{4}{c}{Noise} \vline &\multicolumn{2}{c}{Density} \vline &\multirow{2}{*}{Average} &\multirow{2}{*}{Clean}&\multirow{2}{*}{Noise}&\multirow{2}{*}{Average}\\
~ & None & 0.12$\%$ & 0.6$\%$ &1.2$\%$  &Striped & Gradient  & ~ & None & 0.12$\%$ & 0.6$\%$ &1.2$\%$  &Striped & Gradient&~&~&~\\
\midrule
Jet~\cite{cazals2005estimating}&12.35&12.84&18.33&27.68&13.39&13.13&16.29&20.11	&20.57&31.34&45.19&18.82&18.69&25.79&15.17&15.59&15.38\\
PCA~\cite{hoppe1992surface}&12.29&12.87&18.38&27.52&13.66&12.81&16.25&19.90&20.60&31.33&45.00&19.84&18.54&25.87&15.93&16.32&16.12\\
\midrule
PCPNet~\cite{guerrero2018pcpnet}&9.64&11.51&18.27&22.84&11.73&13.46&14.58&18.47&21.07&32.60&39.93&18.14&19.50&24.95&20.86&21.40&21.13\\
Zhou~\etal~\cite{zhou2020normal}&8.67&10.49&17.62&24.14&10.29&10.66&13.62&-&-&-&-&-&-&-&-&-&-\\
Nesti-Net~\cite{ben2019nesti}&7.06&10.24&17.77&22.31&8.64&8.95&12.49&11.60&16.80&31.61&39.22&12.33&11.77&20.55&13.01&15.19&14.10\\
Lenssen~\etal~\cite{lenssen2020deep}&6.72&9.95&17.18&21.96&7.73&7.51&11.84&11.62&16.97&30.62&39.43&11.21&10.76&20.10&10.24&13.00&11.62\\
DeepFit~\cite{ben2020deepfit}&6.51&9.21&16.73&23.12&7.92&7.31&11.80&11.21&16.39&29.84&39.95&11.84&10.54&19.96&10.33&13.07&11.70\\
Refine-Net~\cite{zhou2022refine}&5.92&9.04&16.52&22.19&7.70&7.20&11.43&-&-&-&-&-&-&-&-&-&-\\
Zhang~\etal~\cite{zhang2022geometry}&5.62&9.91&16.78&22.93&6.68&6.29&11.25&-&-&-&-&-&-&-&-&-&-\\
AdaFit~\cite{li2022graphfit}&5.19&9.05&16.45&21.94&6.01&5.90&10.76&9.09&15.78&29.78&38.74&8.52&8.57&18.41&8.39&12.85&10.62\\
GraphFit~\cite{li2022graphfit}&5.21&8.96&16.12&21.71&6.30&5.86&10.69&8.91&15.73&29.37&38.67&9.10&8.62&18.40&8.39&12.85&10.62\\
Hsurf~\cite{li2022hsurf}&4.17&8.78&16.25&21.61&4.98&4.86&10.11&7.59&15.64&29.43&38.54&7.63&7.40&17.70&7.55&12.33&9.89\\
SHSNet~\cite{li2023shs} &3.95&8.55&16.13&21.53&4.91&4.67&9.96&7.41&15.34&29.33&38.56&7.74&7.28&17.61&7.93&12.40&10.17\\
Li~\etal~\cite{li2023neural}&4.06&8.70&16.12&21.65&4.80&4.56&9.89&7.25&15.60&29.35&38.74&7.60&7.20&17.62&-&-&-\\
CMG-Net~\cite{wu2023cmg} &3.86&8.45&\textbf{16.08}&21.89&4.85&4.45&9.93&7.07&\textbf{14.83}&\textbf{29.04}&38.93&7.43&7.02&17.39&7.64&11.82&9.73\\
MSECNet~\cite{xiu2023msecnet}& 3.84&8.74&16.10&\textbf{21.05}&4.34&4.51&9.76&6.73&15.52&29.19&\textbf{38.06}&\textbf{6.68}&6.70&\textbf{17.15}&6.94&11.66&9.30\\
\midrule
\rowcolor{gray!40}Ours(S)& 3.43&8.38&16.27&21.59&4.18&\textbf{4.10}&9.66&6.71&15.07&29.37&39.30&6.89&6.65&17.33&7.42&11.70&9.56\\
\rowcolor{gray!40}Ours(N)&\textbf{3.42}&\textbf{8.32}&16.22&21.71&\textbf{4.10}&4.14&\textbf{9.65}&\textbf{6.62}&15.03&29.25&38.85&6.86&\textbf{6.63}&17.21&7.42&11.81&9.61\\
\bottomrule
\end{tabular}
\label{tab:1}
\end{table*}

\begin{figure*}[htp!]
    \centering
    \includegraphics[width=0.8\textwidth]{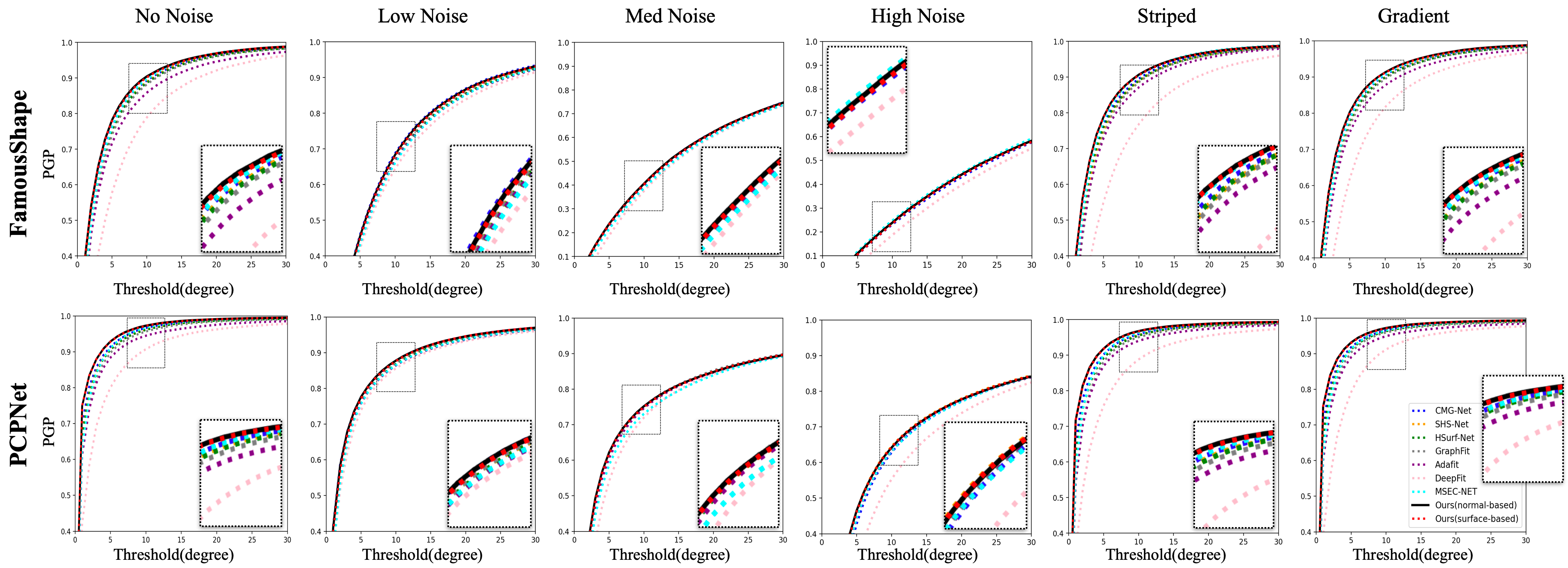}
    \caption{AUC on the PCPNet and FamousShape dataset. X and Y axes are the angle threshold and the percentage of good point (PGP) normals.}
    \label{fig:4}
\end{figure*}

\begin{figure*}[htp]
    \centering
    \includegraphics[width=0.8\textwidth]{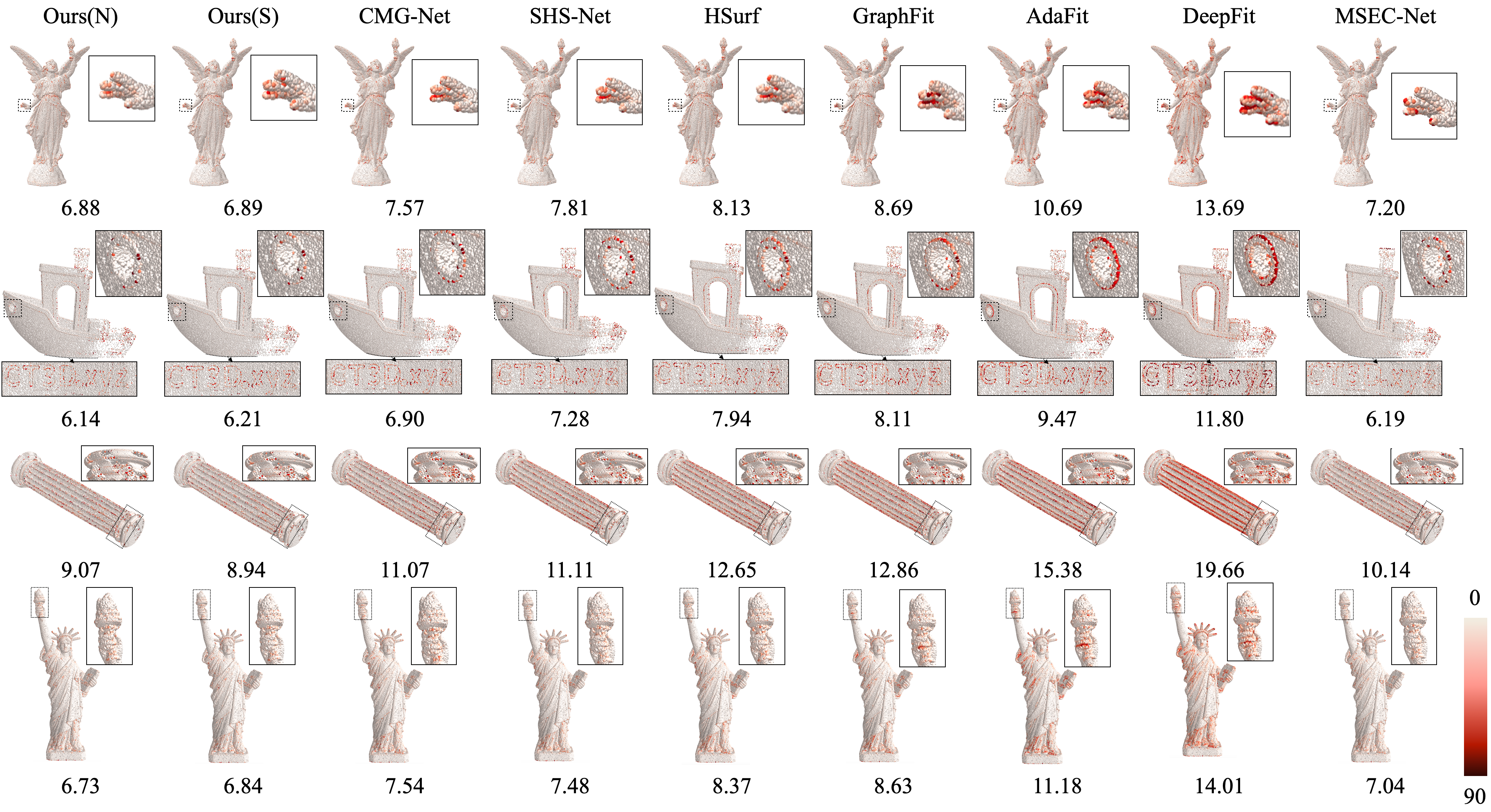}
    \caption{Visualization comparisons on the PCPNet and FamousShape datasets, with numbers indicating RMSEs. The angle error is visualized using a heatmap.}
    \label{fig:5}
\end{figure*}

\begin{figure}[htp!]
    \centering
    \includegraphics[width=0.96\linewidth]{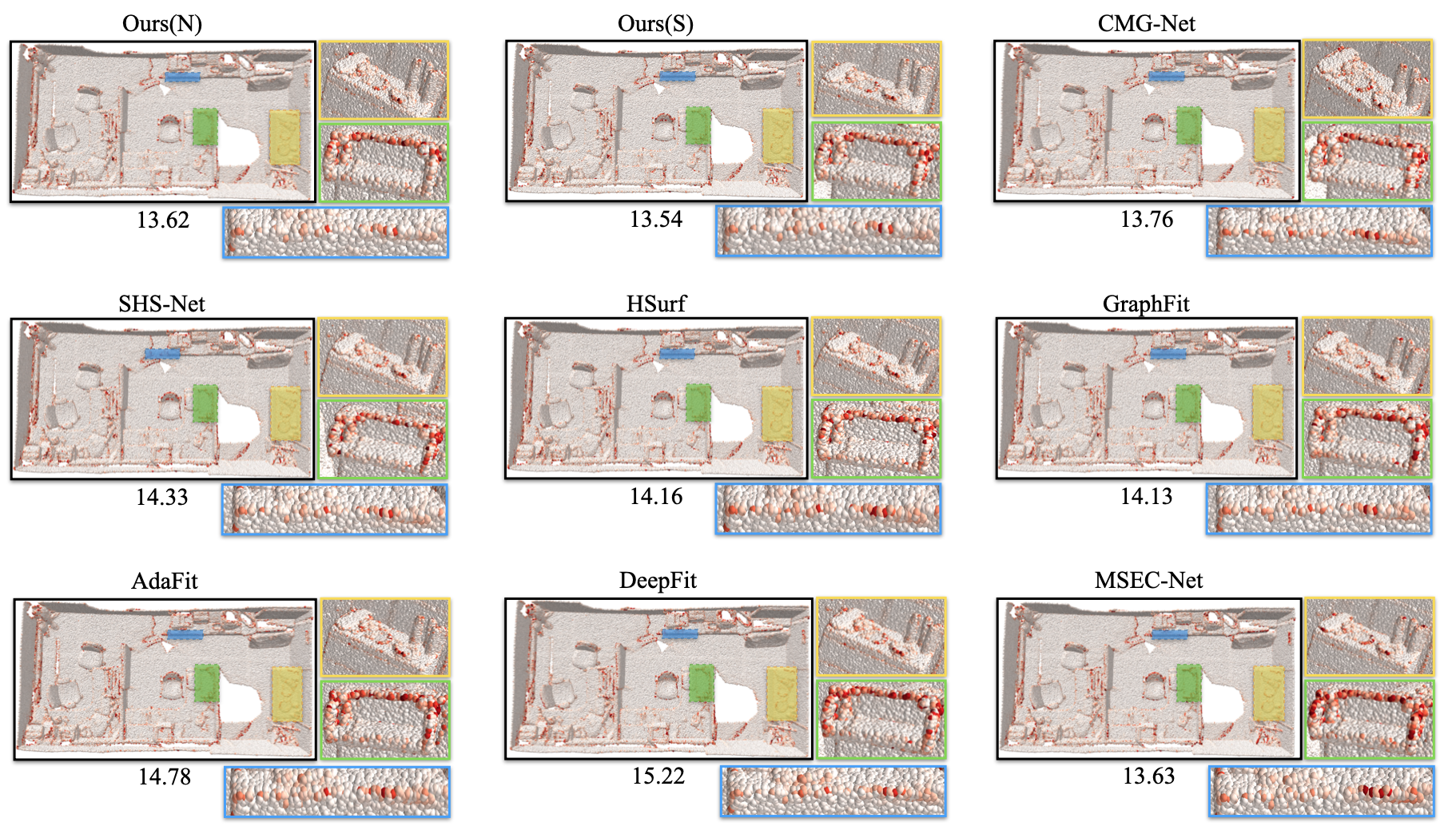}
    \caption{Qualitative comparisons on the SceneNN datasets (Noise: $\sigma$ = 0.3$\%$).}
    \label{fig:6}
\end{figure}

\section{Experiments}
\subsection{Datasets and Settings}
In this study, all comparative experimental models are trained on the same training set from PCPNet dataset~\cite{guerrero2018pcpnet}, which provides ground-truth normals with consistent orientation. Various scales of noise and distribution density are applied during training. To evaluate our method's generalization capability, we utilize both synthetic and real datasets. The synthetic datasets comprise the PCPNet~\cite{guerrero2018pcpnet} and Famous datasets~\cite{erler2020points2surf}, while the real datasets comprise Semantic3D~\cite{hackel2017semantic3d} and SceneNN~\cite{hua2016scenenn}. The test dataset configuration is identical to that of SHS-Net~\cite{li2023shs}. Similar to SHS-Net~\cite{li2023shs}, we employ two metrics, namely, the angle Root Mean Squared Error (RMSE) and the Area Under the Curve (AUC), to quantitatively compare different methods. In the training stage, we use a batch size of 145 and epochs of 800, the Adam optimizer, and a base learning rate of 0.0009. Our network is trained on a single NVIDIA RTX 3090 GPU. For network configuration, in local patch encoding, we randomly select a query point and its 700 neighboring points to create a patch. In global shape encoding, we sample 1200 points from the shape point cloud. Further details regarding parameter settings are provided in the ablation study subsection.

\begin{table}[htp]
\caption{Comparison of the RMSE angle error for corrected unoriented normal estimation of our method to classical geometric methods, and deep learning methods on modified datasets PCPNet.$*$ means the code is uncompleted.} 
\tiny
\centering
\begin{tabular}{l||cccc|cc|c}
\toprule
\multirow{2}{*}{\textbf{Category}} & \multicolumn{4}{c}{Noise} \vline &\multicolumn{2}{c}{Density} \vline &\multirow{2}{*}{Average} \\
~ & None & 0.12$\%$ & 0.6$\%$ &1.2$\%$  &Striped & Gradient \\
\midrule
PCPNet~\cite{guerrero2018pcpnet}&9.62&11.23&17.28&20.16&11.15&11.69&13.52\\
Nesti-Net~\cite{ben2019nesti}&8.43&10.57&15.00&18.16&10.20&10.66&12.17\\
DeepFit~\cite{ben2020deepfit}&6.51&8.98&13.98&19.00&7.93&7.31&10.62\\
AdaFit~\cite{li2022graphfit}&5.21&8.79&13.55&17.31&6.01&5.90&9.46\\
GraphFit~\cite{li2022graphfit}&4.49&8.43&13.00&16.93&5.40&5.20&8.91\\
Hsurf~\cite{li2022hsurf}&4.17&8.52&13.23&16.72&4.98&4.86&8.75\\
SHS-Net~\cite{li2023shs}&3.95&8.29&13.13&16.60&4.91&4.67&8.59\\
CMG-Net~\cite{wu2023cmg}&3.86&8.13&12.55&16.23&4.85&4.45&8.35\\
\midrule
\rowcolor{gray!40}Ours(S)&\textbf{3.43}&8.06&13.21&16.63&\textbf{4.10}&4.18&8.27\\
\rowcolor{gray!40}Ours(N)&3.51&\textbf{7.79}&12.92&\textbf{16.12}&4.32&\textbf{4.05}&\textbf{8.15}\\
\bottomrule
\end{tabular}
\label{tab:3}
\end{table}

\subsection{Normal estimation performance}
\noindent\textbf{Results on synthetic datasets}  As shown in Table~\ref{tab:1}, our method is competitive with existing deep learning-based and traditional methods on synthetic datasets PCPNet~\cite{guerrero2018pcpnet} and Famous datasets~\cite{erler2020points2surf}. Here, 'S' represents the surface-inclusion-based strategy, and 'N' represents the normal-discrepancy-based strategy. Notably, our method achieves superior scores for point clouds with lower noise and varying density. In higher noise scenarios, our method achieves accuracy comparable to mainstream methods. This suggests that our model and sample selection strategy enhance the exploration of local patch patterns, thereby reducing interference from noisy samples during training and leading to more accurate local normal estimation. Subsequently, we compared the accuracy of the most state-of-the-art algorithms on corrected normals sourced from the PCPNet dataset, as depicted in Table~\ref{tab:2}. Our method remains the most competitive on point clouds with high noise, demonstrating the robustness of our normal estimation architecture. Additionally, the AUC results of all methods are shown in Fig.~\ref{fig:4}, where our method demonstrates superior performance, showcasing remarkable stability across various angular thresholds. In Fig.~\ref{fig:5}, qualitative comparison results are presented using a heatmap to illustrate the angular error at each point of the point cloud. Our method exhibits the smallest errors in regions characterized by varying density, intricate geometry, and local details. In Tab.~\ref{tab:2},  we present quantitative comparison results of oriented normal estimation on the PCPNet and FamousShape datasets.  Our method provides the most accurate normals under almost all noise levels and density variations for both datasets. The experimental results demonstrate a significant improvement in normal estimation accuracy by combining the regression network with NeuralGF~\cite{li2024neuralgf}, a global-oriented normal estimation method. This approach contrasts with SHSNet~\cite{li2023shs}, which employs a single network for both normal and orientation estimation.

~

\begin{figure*}[ht!]
    \centering
    \includegraphics[width=0.8\textwidth]{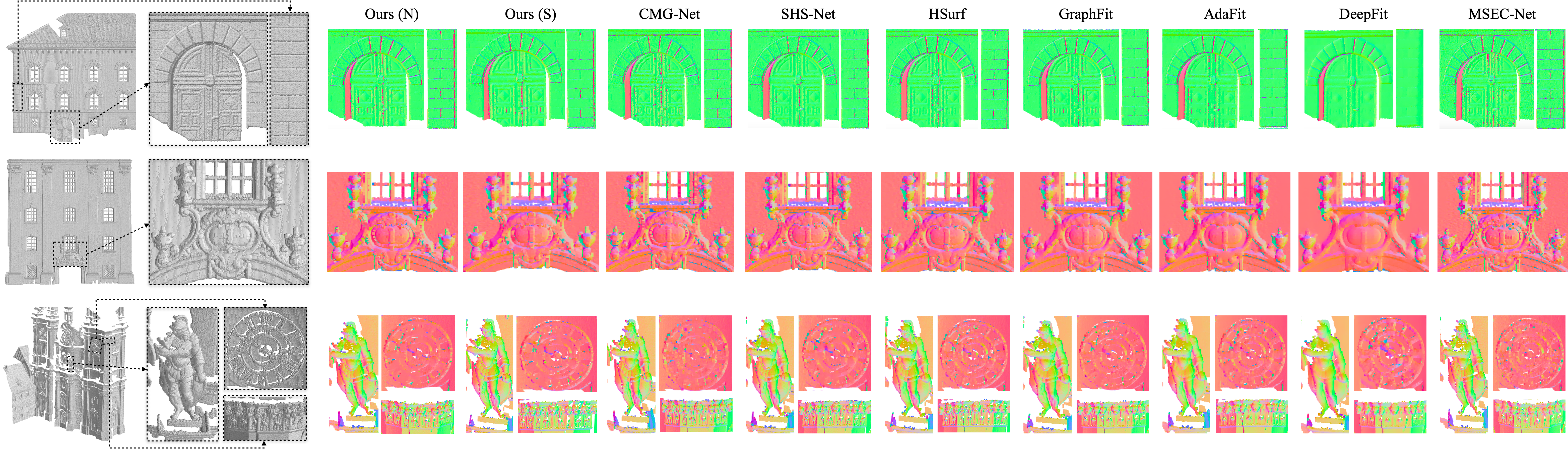}
    \caption{Qualitative comparisons on the Semantic3D dataset, with point normals represented as RGB colors.}
    \label{fig:7}
\end{figure*}

\noindent\textbf{Results on real datasets}
Subsequently, to further evaluate the generalization capacity of our method, we conducted quantitative and qualitative experiments on real datasets, including the indoor SceneNN~\cite{hua2016scenenn} dataset and the outdoor Semantic3D~\cite{hackel2017semantic3d} dataset. The SceneNN dataset is captured using a depth camera, and the ground-truth normals are calculated from the reconstructed meshes. Testing the same models as SHS-Net~\cite{li2023shs}, we present the RMSE scores of different methods in Table~\ref{tab:1}. Our method shows significant improvement over SHS-Net~\cite{li2023shs} and remains competitive with the latest methods, including CMG-Net~\cite{wu2023cmg} and MESCNet~\cite{xiu2023msecnet}. This further demonstrates the effectiveness of our two sample selection strategies proposed in this paper in enhancing the model's generalization capability. 
Additionally, as shown in Figure~\ref{fig:6}, our visualizations illustrate the results of angle error. From these visualized results, it is evident that our method achieves more accurate precision in local details. We also evaluate our method on the Semantic3D dataset~\cite{hackel2017semantic3d}, which comprises laser-scanned point clouds without ground-truth normals. As shown in Figure~\ref{fig:7}, we provide visual comparisons by mapping the normals to the RGB space. It is evident that our method achieves more accurate estimations in the detailed regions, resulting in sharper estimation results. 

\begin{table*}[htp]
\caption{Comparison of the RMSE angle error for oriented normal of our method to other methods on datasets PCPNet and FamousShape.$*$ means the code is uncompleted.} 
\tiny
\centering
\begin{tabular}{l||cccc|cc|c||cccc|cc|c}
\toprule
\multirow{3}{*}{\textbf{Category}}  & \multicolumn{7}{c||}{\textbf{PCPNet Dataset}} & \multicolumn{7}{c}{\textbf{FamousShape Dataset}} \\
\cmidrule(r){2-15}
~ & \multicolumn{4}{c}{Noise} \vline &\multicolumn{2}{c}{Density} \vline &\multirow{2}{*}{Average} & \multicolumn{4}{c}{Noise} \vline &\multicolumn{2}{c}{Density} \vline &\multirow{2}{*}{Average}\\
~ & None & 0.12$\%$ & 0.6$\%$ &1.2$\%$  &Striped & Gradient  & ~ & None & 0.12$\%$ & 0.6$\%$ &1.2$\%$  &Striped & Gradient\\
\midrule
PCA$+$MST~\cite{hoppe1992surface}&19.05&30.20&31.76&39.64&27.11&23.38&28.52&35.88&41.67&38.09&60.16&31.69&35.40&40.48\\
PCA$+$SNO~\cite{schertler2017towards}&18.55&21.61&30.94&39.54&23.00&25.46&26.52&32.25&39.39&41.80&61.91&36.69&35.82&41.31\\
PCA$+$ODP~\cite{metzer2021orienting}&28.96&25.86&34.91&51.52&28.70&32.00&32.16&30.47&31.29&41.65&84.00&39.41&30.72&42.92\\
AdaFit~\cite{zhu2021adafit}$+$MST&27.67&43.69&48.83&54.39&36.18&40.46&41.87&43.12&39.33&62.28&60.27&45.58&42.00&48.76\\
AdaFit~\cite{zhu2021adafit}$+$SNO&26.41&24.17&40.31&48.76&27.74&31.56&33.16&27.55&37.60&69.56&62.77&27.86&29.19&42.42\\
AdaFit~\cite{zhu2021adafit}$+$ODP&26.37&24.86&35.44&51.88&26.45&20.57&30.93&41.75&39.19&44.31&72.91&45.09&42.37&47.60\\
Hsurf-Net~\cite{li2022hsurf}$+$MST&29.82&44.49&50.47&55.47&40.54&43.15&43.99&54.02&42.67&68.37&65.91&52.52&53.96&56.24\\
Hsurf-Net~\cite{li2022hsurf}$+$SNO&30.34&32.34&44.08&51.71&33.46&40.49&38.74&41.62&41.06&67.41&62.04&45.59&43.83&50.26\\
Hsurf-Net~\cite{li2022hsurf}$+$ODP&26.91&24.85&35.87&51.75&26.91&20.16&31.07&43.77&43.74&46.91&72.70&45.09&43.98&49.37\\
\midrule
PCPNet~\cite{guerrero2018pcpnet}&33.34&34.22&40.54&44.46&37.95&35.44&37.66&40.51&41.09&46.67&54.36&40.54&44.26&44.57\\
DPGO$^{*}$~\cite{wang2022deep}&23.79&25.19&35.66&43.89&28.99&29.33&31.14&-&-&-&-&-&-&-\\
SHSNet~\cite{li2023shs} &10.28&13.23&25.40&35.51&16.40&17.92&19.79&21.63&25.96&41.14&52.67&26.39&28.97&32.79\\
Li~\etal~\cite{li2023neural}&12.52&12.97&25.94&\textbf{33.25}&16.81&9.47&18.49&13.22&18.66&39.70&51.96&31.32&11.30&27.69\\
NeuralGF~\cite{li2024neuralgf}&10.60&18.30&24.76&33.45&12.27&12.85&18.70&16.57&19.28&\textbf{36.22}&\textbf{50.27}&17.23&17.38&26.16\\
\midrule
\rowcolor{gray!40}Ours(S)&\textbf{6.67}&\textbf{11.51}&24.69&33.76&8.18&8.66&15.58 &10.52&18.67&36.65&52.68&\textbf{11.07}&10.37&23.33\\
\rowcolor{gray!40}Ours(N)&6.76&11.53&\textbf{24.64}&33.54&\textbf{7.91}&\textbf{8.57}&\textbf{15.49}&\textbf{10.37}&\textbf{18.41}&36.43&51.31& 11.13&\textbf{10.29}&\textbf{22.99}\\
\bottomrule
\end{tabular}
\label{tab:2}
\end{table*}

\begin{table*}[htp]
\caption{Ablation studies for unoriented normals on the PCPNet dataset with the different settings. Here (A) is the single branch with local patch and (B) means two branches with global and local patch.} 
\tiny
\centering
\begin{tabular}{l l|cccc|cc|c| c}
\toprule
~& \multirow{2}{*}{\textbf{Ablation}} & \multicolumn{4}{c}{Noise} \vline &\multicolumn{2}{c}{Density} \vline &\multirow{2}{*}{Average} & \multirow{2}{*}{$\Delta (\%)$}\\
~ &~ & None & 0.12$\%$ & 0.6$\%$ &1.2$\%$  &Striped & Gradient &~ \\
\midrule
\multirow{7}{*}{\textbf{(A)}}&baseline (center loss $+$ neighbor loss)&3.71&8.66&16.13&21.65&4.69&4.48&9.89&-\\
~&baseline $+$ qstn&3.60&8.56&16.61&21.58&4.41&4.73&9.84&0.51\\
~&baseline$+$ qstn $+$ z-loss&3.67&8.63&16.26&21.69&4.52&4.56&9.89&0.00\\
~&best (baseline$+$ qstn $+$ z-loss $+$ neighbor-sin)&3.70&8.50&16.24&21.54&4.51&4.49&9.83&0.61\\
~& best $+$ train with modified gt &3.69&8.49&16.08&21.87&4.73&4.33&9.86&0.30\\
\rowcolor{gray!40}~& best $+$ surface confidence ($\sigma=0.05$) &3.50&8.51&16.26&21.67&4.29&4.16&9.73&1.62\\
\rowcolor{gray!40}~& best $+$ normal confidence ($\sigma=0.05$) &3.44&8.43&16.18&21.98&4.10&4.22&9.72&1.72\\
\midrule
\midrule
\multirow{7}{*}{\textbf{(B)}}&baseline (center loss $+$ neighbor loss)&3.87&8.56&16.44&21.56&4.57&4.84&9.97&-\\
~&baseline $+$ qstn&4.07&8.55&16.50&21.68&4.71&4.90&10.07&-1.00\\
~&baseline$+$ qstn $+$ z-loss&3.53&8.54&16.32&21.58&4.31&4.61&9.82&1.50\\
~&best (baseline$+$ qstn $+$ z-loss $+$ neighbor-sin)&3.52&8.38&16.25&21.45&4.31&4.41&9.72&2.51\\
~& best $+$ train with modified gt &3.62&8.39&16.22&21.92&4.58&4.52&9.87&1.00\\
\rowcolor{gray!40}~& best $+$ surface confidence ($\sigma=0.05$) &3.43&8.38&16.27&21.59&4.18&4.10&9.66&3.11\\
\rowcolor{gray!40}~& best $+$ normal confidence ($\sigma=0.05$) &3.51&8.32&16.31&21.67&4.32&4.05&9.70&2.71\\
\bottomrule
\end{tabular}
\label{tab:4}
\end{table*}

\begin{table*}[htp]
\caption{Ablation studies for unoriented normals on the PCPNet dataset with the different settings. Here (A) is the single branch with local patch and (B) means two branches with global and local patch.} 
\tiny
\centering
\begin{tabular}{l|c|c|cccc|cc|c}
\toprule
\multirow{2}{*}{\textbf{$\sigma$}}&\multirow{2}{*}{surface conf}&\multirow{2}{*}{normal conf} & \multicolumn{4}{c}{Noise} \vline &\multicolumn{2}{c}{Density} \vline &\multirow{2}{*}{Average} \\
~ &~ &~& None & 0.12$\%$ & 0.6$\%$ &1.2$\%$  &Striped & Gradient &~ \\
\midrule
\multirow{2}{*}{0.1}&$\checkmark$&~&6.62&9.50&16.94&22.24&7.98&7.25&11.75\\
\cmidrule(r){2-10}
~&~&$\checkmark$&3.43&8.40&16.26&21.72&4.27&4.27&9.72\\
\midrule
\multirow{2}{*}{0.08}&$\checkmark$&~&6.41&9.45&16.69&22.07&7.64&6.92&11.53\\
\cmidrule(r){2-10}
~&~&$\checkmark$&3.43&8.39&16.28&21.67&4.20&4.09&9.68\\
\midrule
\multirow{2}{*}{0.06}&$\checkmark$&~&3.62&8.52&16.28&21.90&4.44&4.32&9.85\\
\cmidrule(r){2-10}
~&~&$\checkmark$&3.42&8.32&16.23&21.71&4.10&4.14&9.65\\
\midrule
\multirow{2}{*}{0.05}&$\checkmark$&~&3.43&8.38&16.27&21.59&4.18&4.10&9.66\\
\cmidrule(r){2-10}
~&~&$\checkmark$&3.51&8.32&16.31&21.67&4.32&4.05&9.70\\
\midrule
\multirow{2}{*}{0.03}&$\checkmark$&~&3.77&8.43&16.44&21.54&4.74&4.57&9.91\\
\cmidrule(r){2-10}
~&~&$\checkmark$&3.60&8.39&16.30&21.59&4.349&4.293&9.75\\
\midrule
\multirow{2}{*}{0.02}&$\checkmark$&~&6.24&9.42&16.78&22.08&7.49&6.93&11.49\\
\cmidrule(r){2-10}
~&~&$\checkmark$&3.52&8.46&16.34&22.14&4.21&4.23&9.82\\
\midrule
\multirow{2}{*}{0.01}&$\checkmark$&~&3.40&8.40&16.34&22.14&4.10&4.24&9.77\\
\cmidrule(r){2-10}
~&~&$\checkmark$&3.44&8.46&16.36&22.12&4.07&4.00&9.74\\
\bottomrule
\end{tabular}
\label{tab:5}
\end{table*}

\begin{figure*}[htp!]
    \centering
    \includegraphics[width=0.8\textwidth]{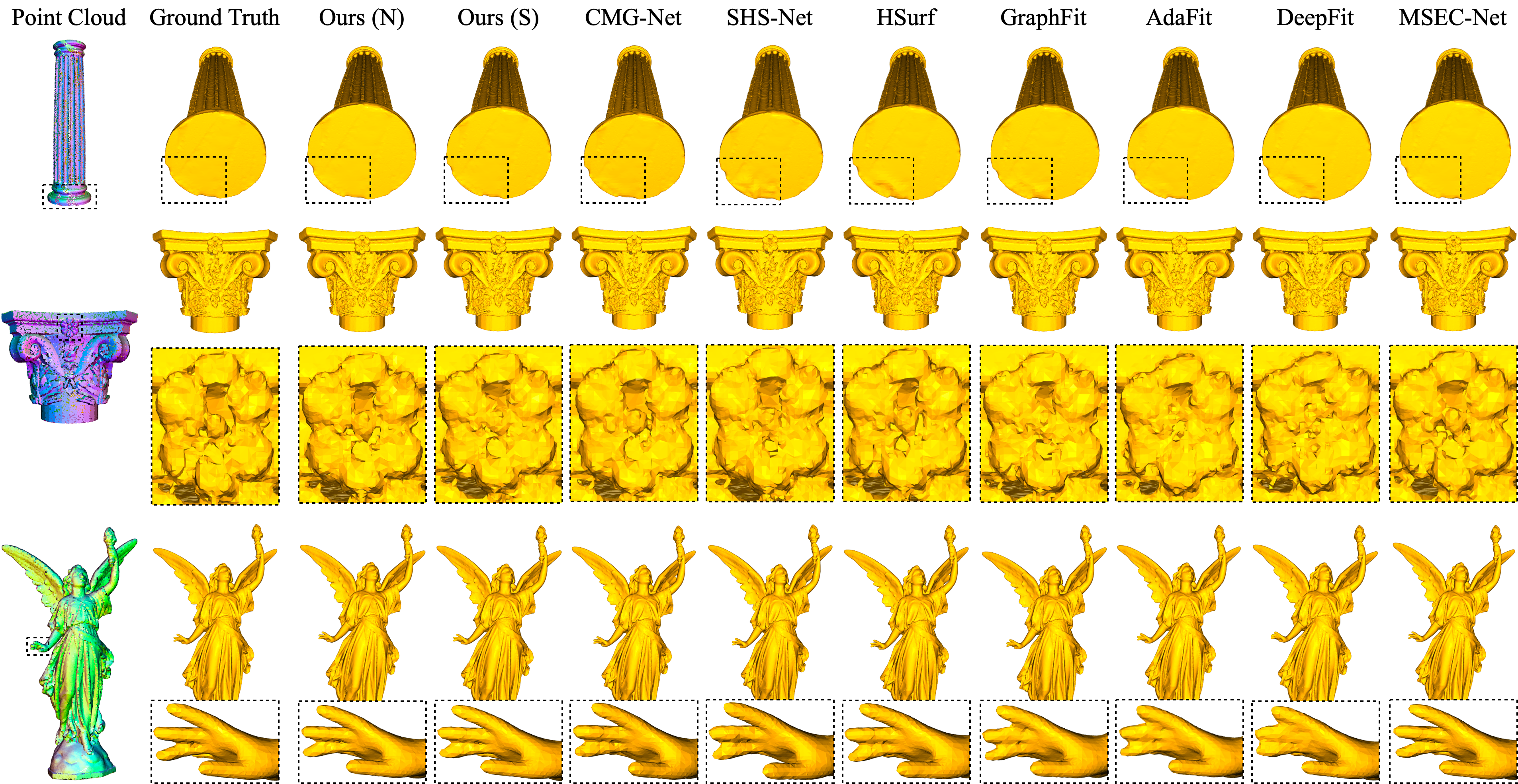}
    \caption{The comparison of Poisson surface reconstruction using normals estimated from different methods.}
    \label{fig:8}
\end{figure*}

\subsection{Ablation Study}
\noindent\textbf{Parameter Setting} Considering that the confidence values (including surface-inclusion-based confidence and normal-discrepancy-based confidence) used in our work will be utilized to suppress unreasonable samples during training, the choice of confidence values will affect model accuracy. The selection of confidence values is influenced by the sigma value, so in this section, we will provide a more detailed discussion on the choice of $\sigma$ values. Considering that the confidence values estimated by our surface-inclusion-based and normal-discrepancy-based strategies play a crucial role in sample selection during training, their choice significantly impacts model accuracy. The selection of these confidence values is influenced by the sigma value. In this section, we will provide a more detailed and insightful discussion on the choice of $\sigma$ values. 
Firstly, if the value of $\sigma$ is set too high, the confidence values among samples become similar, making it difficult to distinguish differences between them. Conversely, if $\sigma$ is too low, many noisy samples may struggle to participate in the training process. Therefore, it is crucial to choose a moderate value. As shown in Table~\ref{tab:5}, a $\sigma$ value of 0.05 is appropriate for calculating surface-inclusion-based confidence, while a $\sigma$ value of 0.06 is most suitable for normal-discrepancy-based confidence.

~

\noindent\textbf{Network Architecture} To demonstrate the effectiveness of our architecture and the reweighting-based sample selection strategy, we conducted extensive ablation experiments on the PCPNet dataset. As shown in Table~\ref{tab:4} (A)-(B), we compared two baselines: a single-branch architecture with local patch input and a two-branch architecture with both global and local patch inputs. It is evident that without the introduction of additional losses and the QSTN module, the inclusion of extra global information can negatively impact model training, leading to decreased model accuracy (the average RMSE decreases from 9.89 to 9.97). Tabl~\ref{tab:4} also shows that, regardless of whether it's a single-branch or a two-branch framework, the introduction of the QSTN module, z-direction transformation loss, and neighborhood consistency loss significantly improves model accuracy (compared to the baseline, they lead to improvements of $0.61\%$  and $2.51\%$, respectively). With the enhancement of model representation and the introduction of effective constraints, the global branch provides valuable information that further improves the performance of normal estimation. Therefore, we ultimately chose the two-branch framework as the final normal estimation model. Finally, we evaluated the effectiveness of the reweighting-based sample selection method, as shown in the last three columns of Tab.~\ref{tab:4} (A) and (B). It can be observed that directly training with the corrected normals, whether in a single-branch or multi-branch architecture, does not lead to significant improvements, as employed in CMG-NET~\cite{wu2023cmg}. In contrast, the loss functions based on surface-inclusion and normal-discrepancy-based confidence suppression proposed in this paper bring more noticeable improvements, especially in the multi-branch architecture where the enhancement is more prominent. This suggests that the corrected normals may not necessarily represent true normals, and a strategy based on reweighting may offer a more flexible constraint. Here, we set the $\sigma$ value to 0.05 in these ablation experiments.

\subsection{Application of the Proposed Method}
\noindent\textbf{Poisson Reconstruction} To explore the potential of our proposed method for other tasks, we investigated the effectiveness of normals in point surface reconstruction. We utilized the classic Poisson reconstruction method~\cite{kazhdan2006poisson} for this purpose. As depicted in Fig.~\ref{fig:8}, we present surfaces reconstructed using normals estimated with various methods. The results indicate that surfaces reconstructed with normals from our method are more accurate and exhibit sharper details and boundaries in local regions.

~

\begin{figure*}[htp!]
    \centering
    \includegraphics[width=0.8\textwidth]{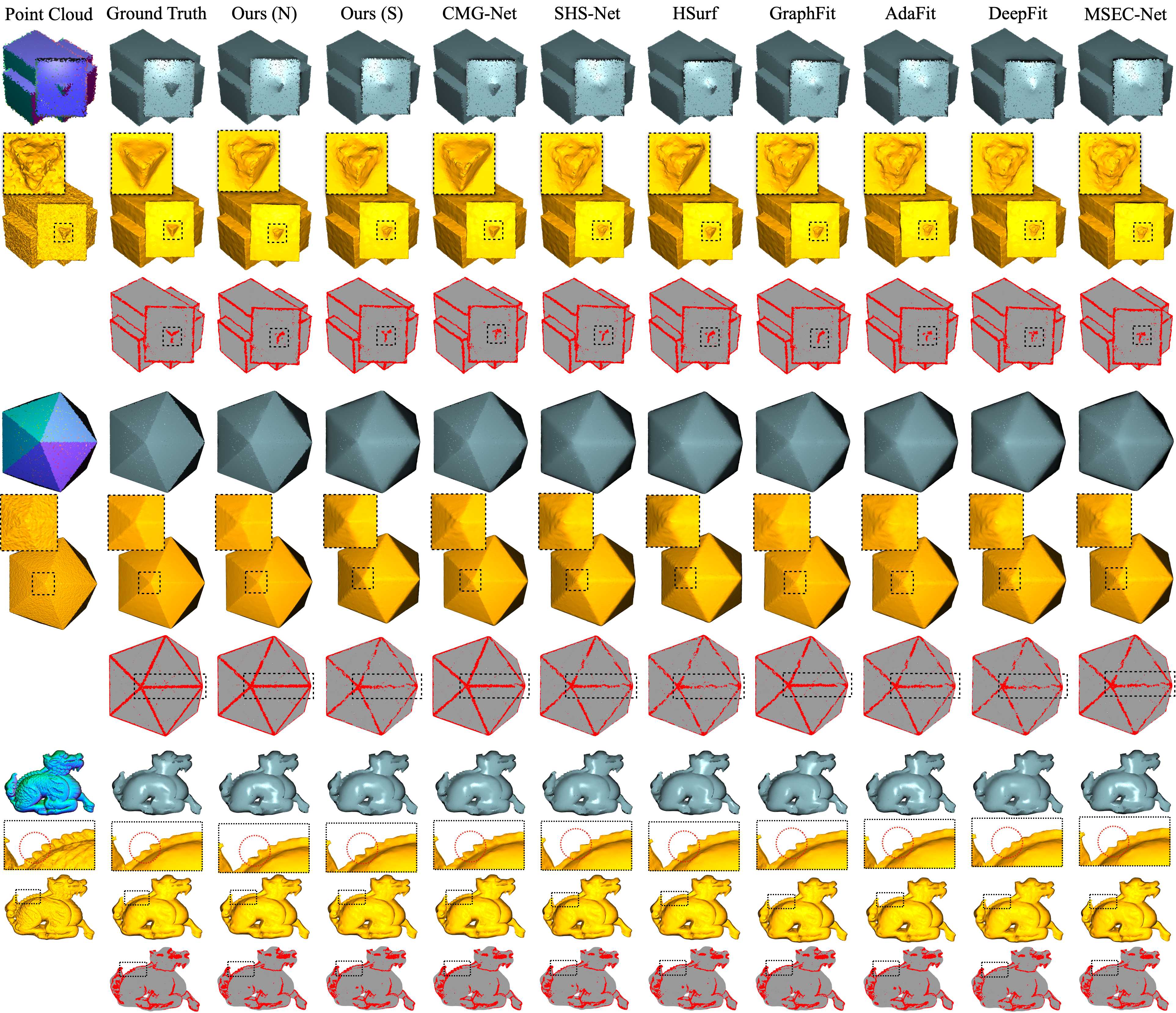}
    \caption{Qualitative results of point cloud denoising. For each instance: the first row displays the denoised point clouds, the second row exhibits the corresponding reconstructed surfaces, and the third row demonstrates feature detection results of the denoised point clouds.}
    \label{fig:9}
\end{figure*}

\noindent\textbf{Point Cloud Denoising} In our study, we also apply a normal-based denoising method~\cite{lu2020low} to validate the accuracy of normal estimation provided by our method. The denoised point clouds, the corresponding reconstructed surfaces, and feature detection results are displayed in Fig.~\ref{fig:9}. The visualization clearly demonstrates that our method's normal estimation significantly contributes to point cloud denoising. In comparison to other methods, ours yields smoother surfaces in flat regions while retaining sharp features at edges. 

\section{Conclusion}
In our study, we introduce a novel framework for normal estimation that integrates normal estimation architecture with orientation algorithms to achieve higher accuracy and consistency. We propose two confidence-based sample selection training strategies to mitigate the impact of corrupted samples on the training model, ensuring its robustness. Extensive experiments validate that our method outperforms competitors in both accuracy and robustness for normal estimation. Additionally, we demonstrate its effectiveness in tasks such as normal-based reconstruction and point cloud denoising.

\bibliographystyle{IEEEtran}
\bibliography{manuscript}

\newpage 
\section*{Biography} 

\vspace{-33pt}
\begin{IEEEbiography}[{\includegraphics[width=1in,height=1.25in,clip,keepaspectratio]{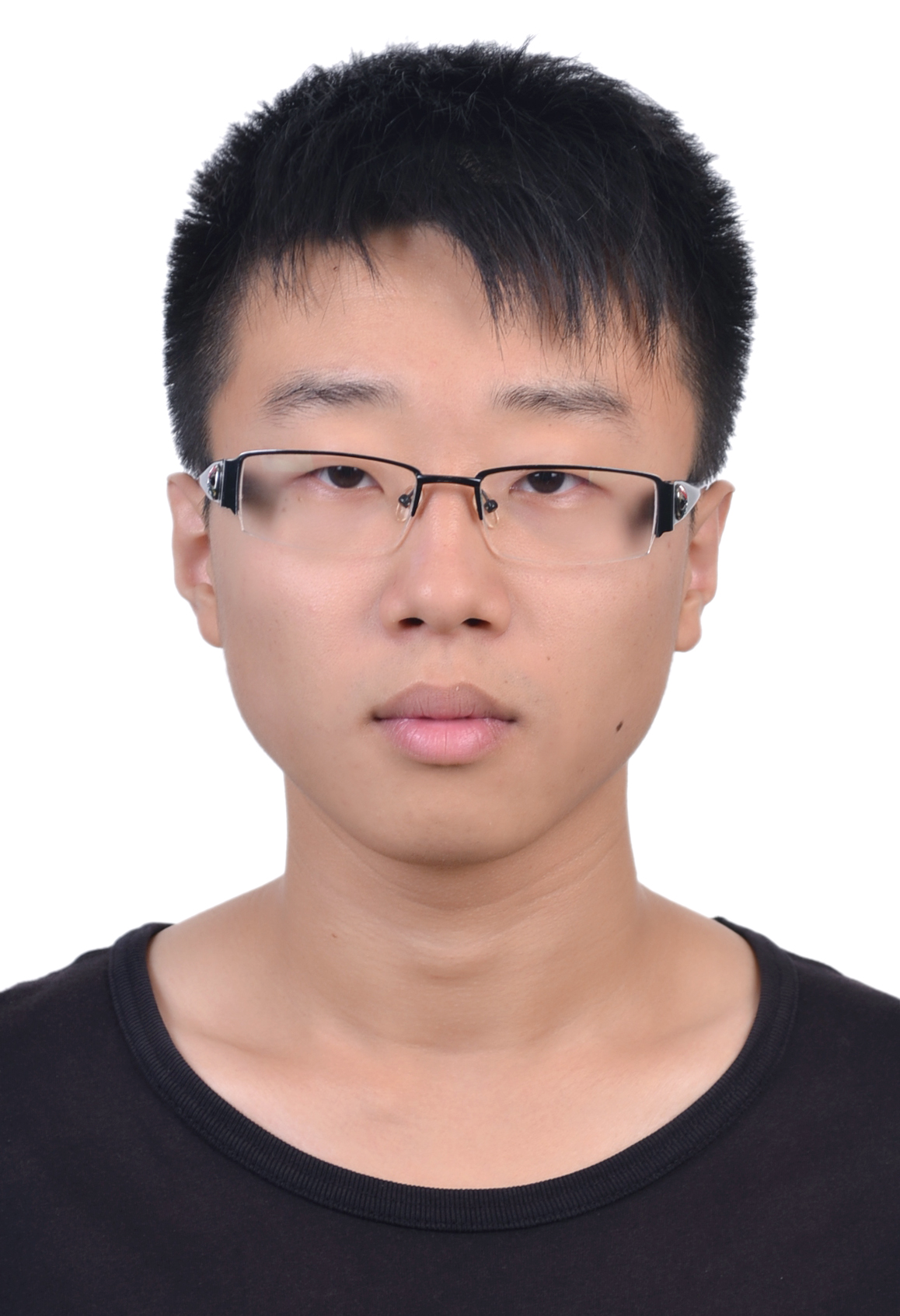}}]{Jun Zhou} received the BSc and PhD degrees in computational mathematics from Dalian University of Technology, China, in 2013 and 2020. He is a lecturer in the College of Information Science and Technology at Dalian Maritime University, China. His research interests include computer graphics, image processing and machine learning.\end{IEEEbiography}

\vspace{-33pt}
\begin{IEEEbiography}[{\includegraphics[width=1in,height=1.25in,clip,keepaspectratio]{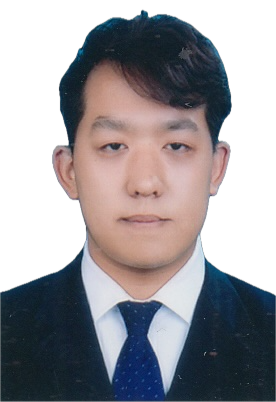}}]{Yaoshun Li} received the bachelor's degree in oil and gas storage and transportation engineering from Liaoning Petrochemical University in 2018. He is currently pursuing a master's degree at Dalian Maritime University. His research interests include computer vision and computer graphics.\end{IEEEbiography}

\vspace{-33pt}
\begin{IEEEbiography}[{\includegraphics[width=1in,height=1.25in,clip,keepaspectratio]{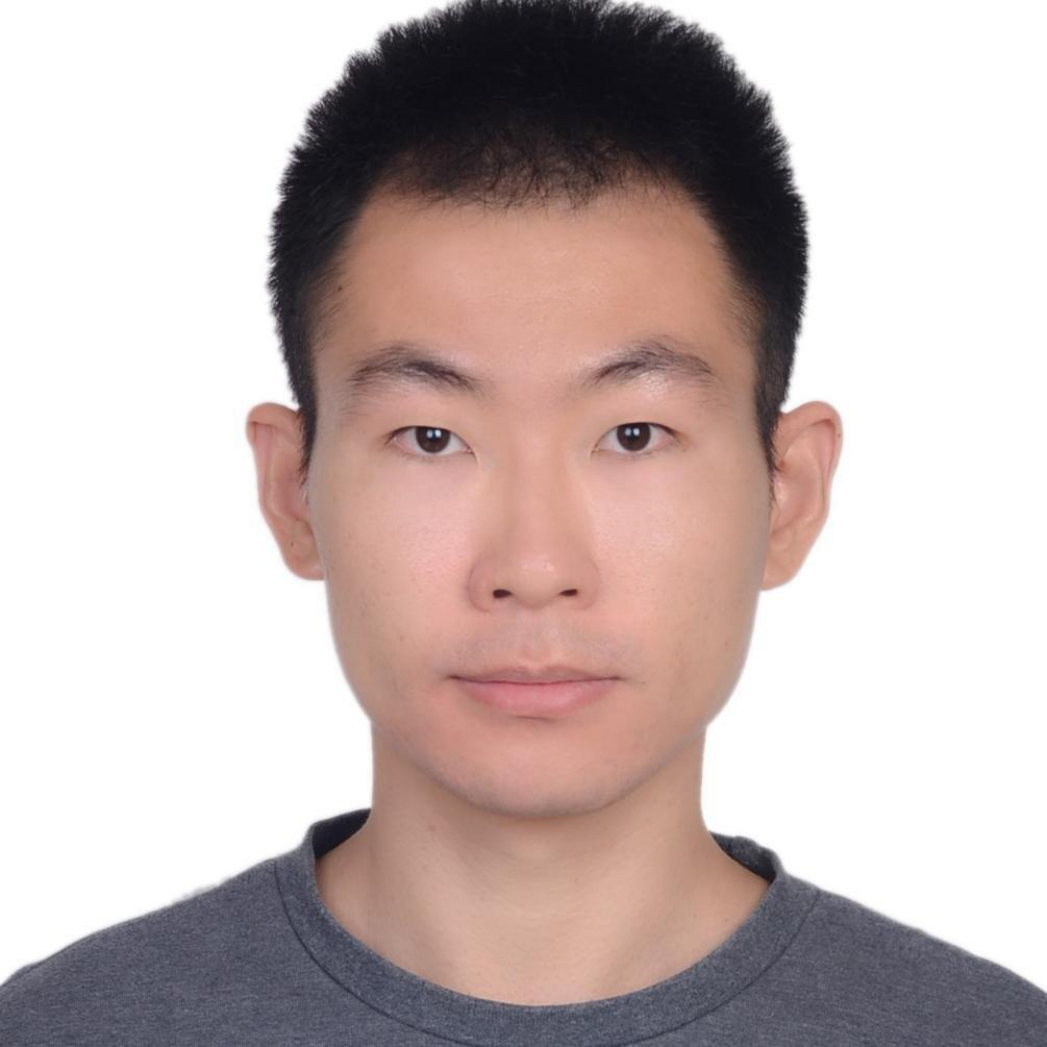}}]{Hongchen Tan} is a  Lecturer  of Artificial Intelligence Research Institute at Beijing University of Technology. He received Ph.D degrees in computational mathematics from  the Dalian University of Technology in 2021. His research interests are  Person Re-identification,  Image Synthesis, and  Referring Segmentation.\end{IEEEbiography}

\vspace{-33pt}
\begin{IEEEbiography}[{\includegraphics[width=1in,height=1.25in,clip,keepaspectratio]{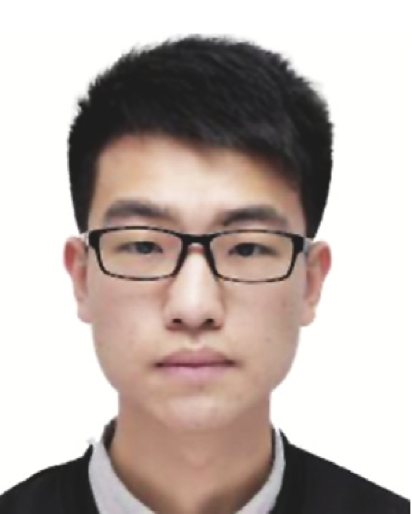}}]{Mingjie Wang} is an associate professor in the School of Science at Zhejiang Sci-Tech University. He received the Ph.D degree from University of Guelph in 2022. His research interests include image processing, computer vision and deep learning.\end{IEEEbiography}

\vspace{-33pt}
\begin{IEEEbiography}[{\includegraphics[width=1in,height=1.25in,clip,keepaspectratio]{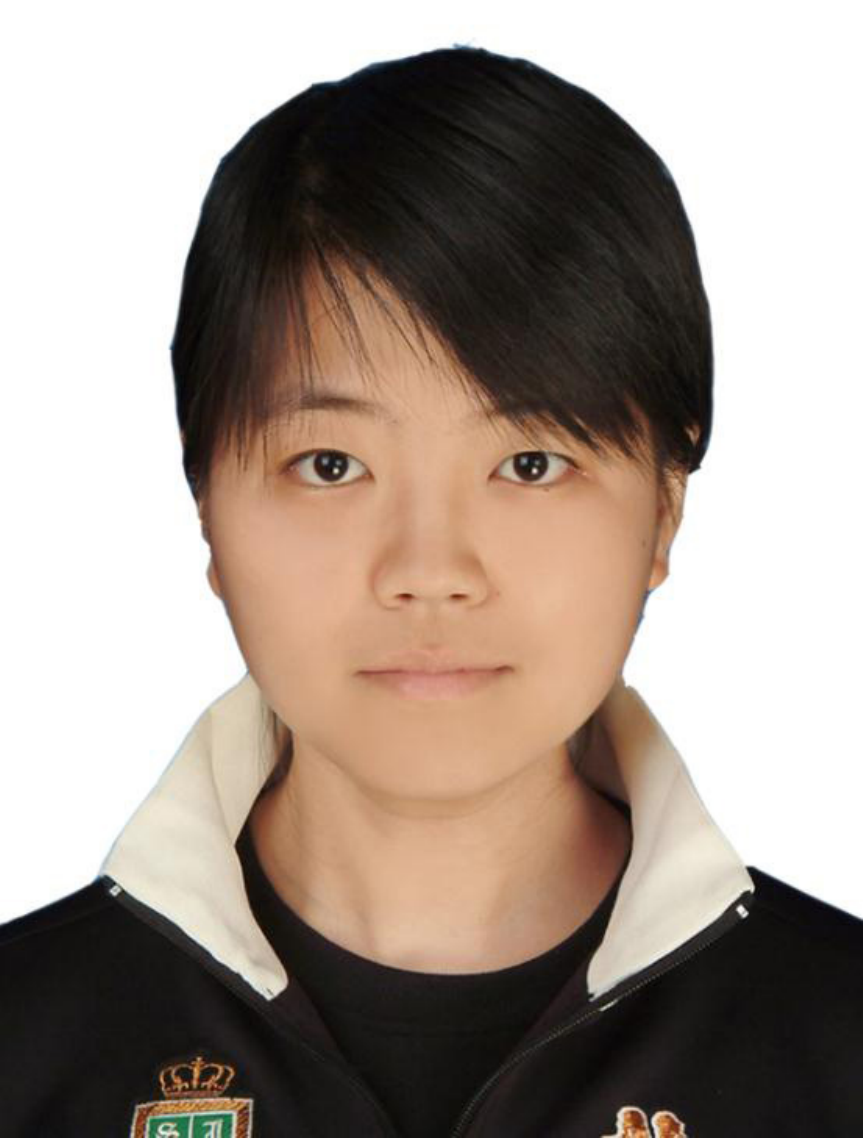}}]{Nannan Li} is an associate professor in School of Information Science and Technology at Dalian  Maritime University. She received her B.S. and M.D-Ph.D  in Computational  Mathematics at Dalian University of Technology. Her  research interests include computer graphics, differential geometry analysis and processing, computer
 aided geometric design and machine learning.\end{IEEEbiography}

\vspace{-33pt}
\begin{IEEEbiography}[{\includegraphics[width=1in,height=1.25in,clip,keepaspectratio]{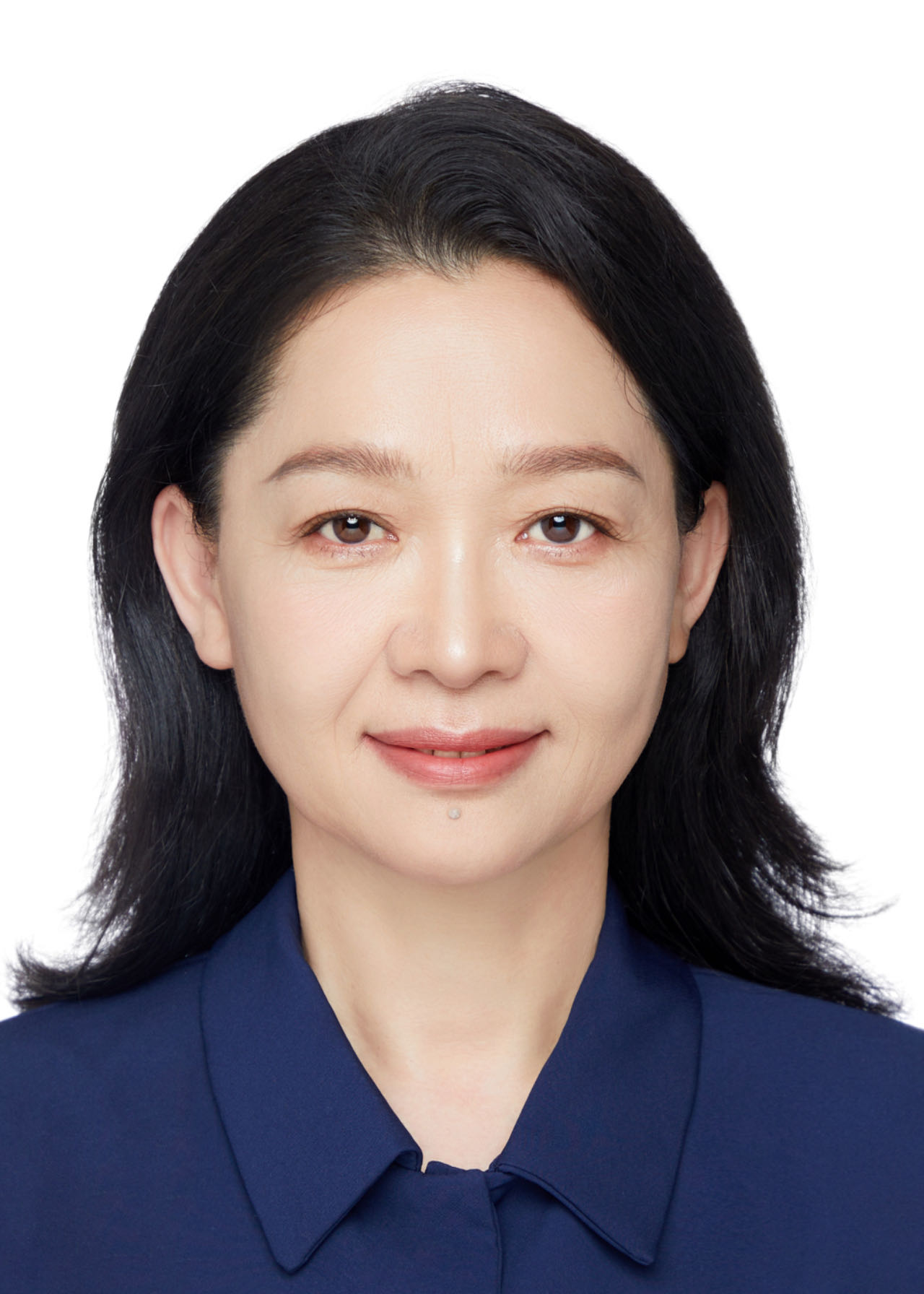}}]{Xiuping Liu} received the BSc degree from Jilin University, China, in 1990, and the PhD degree from the Dalian University of Technology, China, in 1999, respectively. She is a professor with the Dalian University of Technology. Between 1999 and 2001, she conducted research as a postdoctoral scholar in the School of Mathematics, Sun Yat-sen University, China. Her research interests include shape modeling and analyzing.\end{IEEEbiography}

\end{document}